\documentclass{article}

\PassOptionsToPackage{numbers, compress}{natbib}

\usepackage[final,nonatbib]{neurips_2023}

\usepackage{colortbl}
\newcommand{\gray}[0]{\cellcolor[gray]{0.9}}

\usepackage[utf8]{inputenc} 
\usepackage[T1]{fontenc}    
\usepackage[pagebackref,breaklinks,colorlinks]{hyperref}       
\usepackage{url}            
\usepackage{booktabs}       
\usepackage{amsfonts}       
\usepackage{nicefrac}       
\usepackage{microtype}      
\usepackage{xcolor}         
\usepackage{graphicx}
\usepackage{wrapfig}
\usepackage{multirow}
\usepackage{amsmath}        
\usepackage{textcomp, gensymb}
\usepackage[ruled,vlined]{algorithm2e}
\usepackage{enumitem, xspace}
\usepackage{pifont}
\newcommand{\cmark}{\ding{51}}%
\newcommand{\xmark}{\ding{55}}%

\usepackage{subcaption}

\title{Leveraging Vision-Centric Multi-Modal Expertise \\
for 3D Object Detection}

\author{
Linyan Huang$^{1}$\quad
Zhiqi Li$^{2}$\quad Chonghao Sima$^{1}$ \quad 
Wenhai Wang$^{1}$ \\
\textbf{Jingdong Wang}$^{3}$\quad \textbf{Yu Qiao}$^{1}$\quad \textbf{Hongyang Li}$^{1}$  \\[0.2cm]
$^{1}$Shanghai AI Lab\quad $^{2}$Nanjing University \quad $^{3}$Baidu 
}

\begin{document}

\maketitle

\begin{abstract}

Current research is primarily dedicated to advancing the accuracy of camera-only 3D object detectors (apprentice) through the knowledge transferred from LiDAR- or multi-modal-based counterparts (expert).
However, the presence of the domain gap between LiDAR and camera features, coupled with the inherent incompatibility in temporal fusion, significantly hinders the effectiveness of distillation-based enhancements for apprentices.
%
Motivated by the success of uni-modal distillation, an apprentice-friendly expert model would predominantly rely on camera features, while still achieving comparable performance to multi-modal models.
%
%
%
To this end, we introduce \textbf{VCD}, a framework to improve the camera-only apprentice model, including an apprentice-friendly multi-modal expert and temporal-fusion-friendly distillation supervision.
The multi-modal expert \textbf{VCD-E} adopts an identical structure as that of the camera-only apprentice in order to alleviate the feature disparity, and leverages LiDAR input as a depth prior to reconstruct the 3D scene, achieving the performance on par with other heterogeneous multi-modal experts.
Additionally, a fine-grained trajectory-based distillation module is introduced with the purpose of individually rectifying the motion misalignment for each object in the scene.
With those improvements, our camera-only apprentice \textbf{VCD-A} sets new state-of-the-art on nuScenes with a score of 63.1\% NDS.
The code will be released at \url{https://github.com/OpenDriveLab/Birds-eye-view-Perception}.
\end{abstract}

\section{Introduction}

The camera-only 3D perception has garnered increasing attention in autonomous driving perception tasks~\cite{li2022delving, hu2023planning, Tong_2023_ICCV, wang2023openlanev2}.
Although camera-only models possess the advantages of low deployment cost and ease of widespread application, they still fall behind state-of-the-art models that leverage LiDAR sensors regarding perception accuracy.
%
Researchers have recently employed distillation methods to transfer knowledge from a powerful expert model into a camera-only apprentice model, with the expectation of leveraging the expertise of these stronger expert models to enhance the capability of the camera-only models. 
Existing 3D perception distillation methods often adopt expert models with the best performance, such as LiDAR-based models~\cite{yin2021centerpoint} or multi-modal fusion models~\cite{bai2022transfusion,yang2022deepinteraction, li2022uvtr}. 
However, the presence of the domain gap between LiDAR and camera features hampers knowledge transfer during distillation, resulting in limited improvements in practical applications.
An alternative expert model is the large-scale camera-only model~\cite{li2022bevlgkd, Zeng_2023_CVPR}.
Despite eliminating the domain gap between the camera-only expert model and the apprentice model, the expert model falls short in terms of effectiveness due to the inherent lack of precise geometry information. 
Likewise, it fails to yield a satisfactory improvement to the apprentice model.
Hence, a desirable expert should meet two essential requirements: attaining state-of-the-art performance and minimizing the domain gap.

\begin{figure}
  \begin{subfigure}[t]{0.5\textwidth}
    \includegraphics[width=\textwidth]{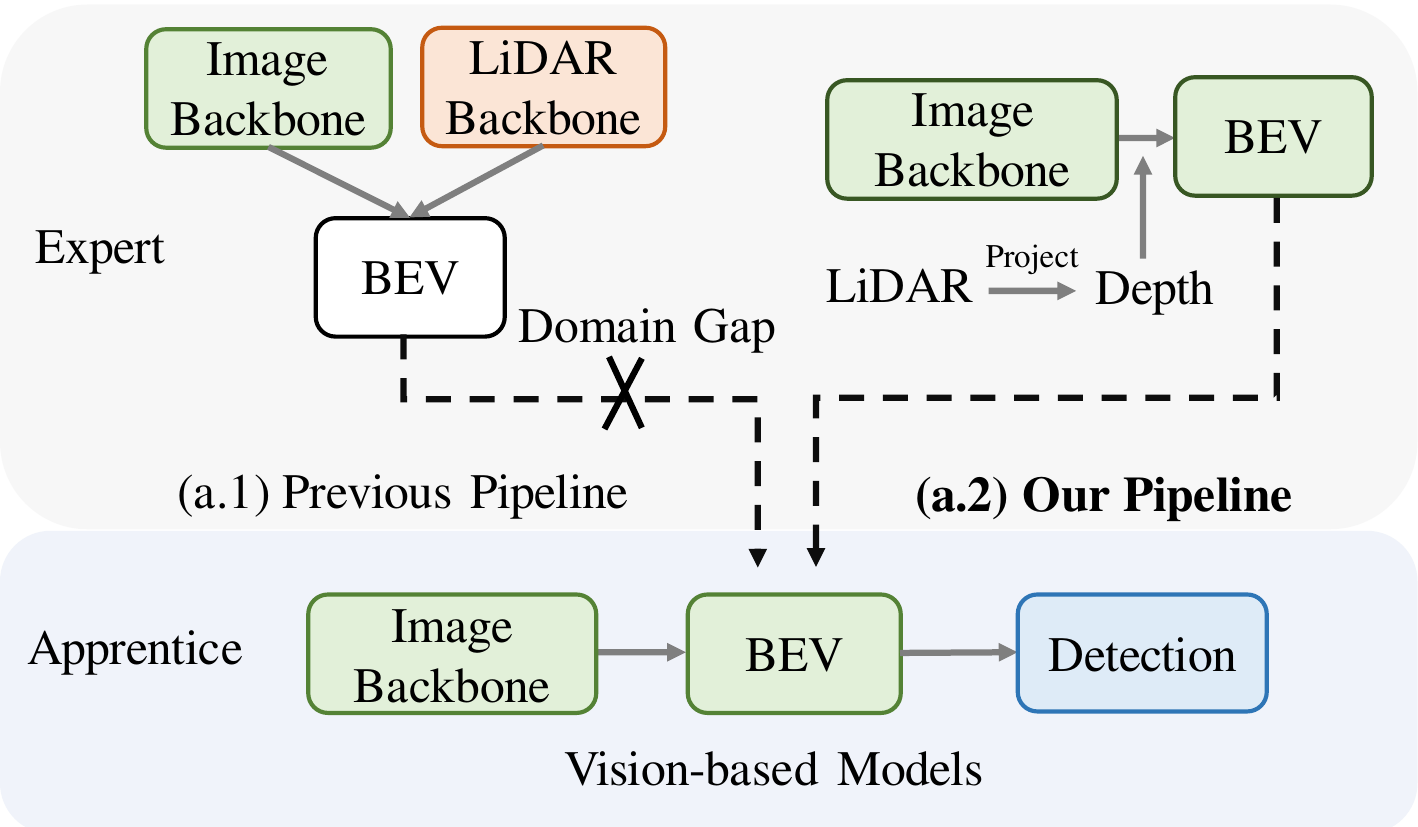}
    \caption{Comparison of various multimodal fusion methods.}
    \label{fig-a}
  \end{subfigure}\hfill
  \begin{subfigure}[t]{0.475\textwidth}
    \includegraphics[width=\textwidth]{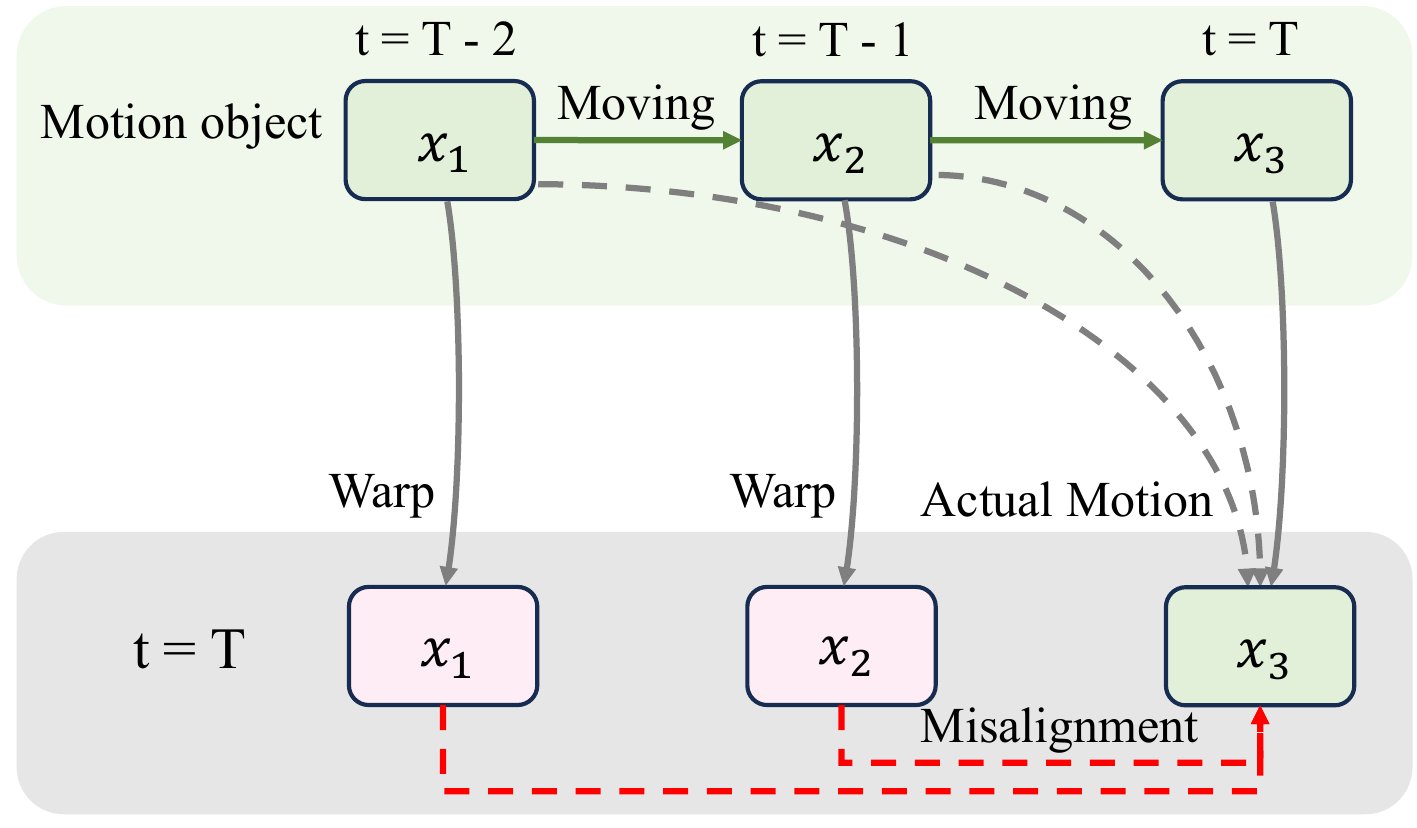}
    \caption{The misalignment of moving objects.}
    \label{fig-b}
  \end{subfigure}
  \caption{(a) The existing pipelines require camera and LiDAR backbones, while our pipeline eliminates the need for LiDAR. Using point cloud depth, we directly transform image features into BEV space to create a vision-centric expert.
  (b) The warping of an object in the historical frame into the current timestamp results in a false position in the current frame due to assuming the object is stationary. The green rectangle represents true positives, while the pink rectangle indicates false positives. $x_i$ denotes the various positions of the object in the historical timestamp.} 
  \label{multi-modal-fusion}
\end{figure}

Furthermore, current distillation methods fall short in compatibility with long-term temporal fusion, which is an essential component in cutting-edge camera-only 3D detectors~\cite{han2023recurrent, park2022solofusion}.
%
Long-term temporal modeling has shown considerable potential in enhancing the accuracy of depth estimation and detection performance, but it introduces the issue of motion misalignment. 
Previous methods for BEV distillation have followed two distinct approaches, either distilling the entire BEV space without sufficient attention to the foreground objects~\cite{romero2014fitnets} or exclusively distilling the foreground object regions~\cite{chen2022bevdistill, huang2022tigbev}, thereby overlooking the motion misalignment issue resulting from the long-term temporal fusion.
As shown in Fig.~\ref{multi-modal-fusion} (b), this misalignment occurs when past scenes are transformed into the current scene coordinates based solely on ego-motion, assuming all objects are stationary. 
While in reality, dynamic objects will cause the misalignment, thus interfering with the temporal fusion features.
This is more challenging in the case of long-term temporal fusion. 
Existing methods, such as StreamPETR~\cite{wang2023streampetr}, introduce LayerNorm~\cite{ba2016layernorm} for dynamic object modeling, but the effects of incorporating velocity and time variables in the model are relatively minor.

To address the aforementioned challenges, we first propose a vision-centric expert, termed as \textbf{VCD-E}, which incorporates LiDAR information to enhance the accuracy of depth input. 
In this context, the term ``vision-centric'' refers to the utilization of prominent features derived from camera input, distinguishing it from approaches that heavily rely on LiDAR-based features.
This model is distinct from conventional multi-modality fusion techniques by eliminating LiDAR backbone.
By solely integrating LiDAR depth and long-term temporal fusion under bird’s-eye-view (BEV), our model achieves comparable performance to state-of-the-art multi-modal fusion methods~\cite{liang2022bevfusion} by only encoding image modality. 
As illustrated in Fig.~\ref{multi-modal-fusion} (a), different from previous fusion methods that adopt two modality-specific backbones, we only leverage a single branch to generate semantic features based on image input, and point clouds only provide depth information. 
Compared to previous fusion methods, our approach eliminates the need for intricate training strategies or specialized fusion module designs while reaching a comparable performance to current fusion methods~\cite{liang2022bevfusion,li2022uvtr}. 
More importantly, the vision-centric multi-modal model has the exact same architecture as the camera-only apprentice model, and the generated BEV spatial representation is solely from image features. 
The domain gap is significantly alleviated through the distillation of knowledge from the proposed multi-modal expert model to the camera-only apprentice model.
Due to the advantageous characteristic of domain consistency, our apprentice model acquires substantial benefits from the expert, surpassing previous distillation methods~\cite{chen2022bevdistill,shu2021cwd}.

%
To mitigate the incompatibility arising from the motion misalignment of dynamic objects, we further propose a trajectory-based distillation module. 
In this paper, our primary focus revolves around foreground objects, while simultaneously incorporating a meticulous consideration of their historical trajectories.
Specifically, by warping dynamic targets from history to the current frame, we derive the motion trajectory associated with each individual object.
Then we use the trajectory of each object to query BEV features of the apprentice and expert models respectively. 
By leveraging the trajectory features of the expert model to optimize the corresponding features of the apprentice model, the latter can acquire the ability to mitigate the interference arising from motion misalignment.
In addition, to enhance the depth perception ability, we diffuse the depth of the foreground part into the 3D space, modeling occupancy to obtain grid-based supervision to assist in depth prediction for objects.


In summary, we propose the multi-modality expert and camera-only apprentice models, termed
as \textbf{VCD-E} and \textbf{VCD-A}, respectively. 
Our contributions are summarized as follows:

 $\bullet$ We construct a vision-centric multi-modal expert that solely encodes the image modality, eliminating the need for a LiDAR backbone.
 For the first time, we demonstrate that the expert can deliver performance on par with other state-of-the-art multi-modal methods while being significantly simpler.
    
 $\bullet$ Due to its homogeneous characteristics and superior performance, the vision-centric expert has been proven effective in distilling knowledge to vision-based models. 
 The effects are significant across a range of model sizes, from compact to more extensive architectures.
 
$\bullet$ We propose trajectory-based distillation and occupancy reconstruction modules, which supervise both static and dynamic objects to alleviate misalignment during long-term temporal fusion. Combined with the constructed expert model, we enhance the performance of the vision-based models and achieve state-of-the-art on the nuScenes val and test leaderboard.

\section{Related Work}
\label{gen_inst}

In this section, we review previous studies in the areas of 3D object detection, multi-modality fusion, knowledge distillation, focusing on the techniques and methods most relevant to our research.

\paragraph{3D Object Detection.}

3D object detection has recently gained significant popularity in the context of autonomous driving and robotics.
Detection methods generally fall into two categories: vision-based 3D object detection ~\cite{zhou2021monocular, Zhou2022monocular, chen2022persformer, li2023voxelformer, wang2021fcos3d, li2022bevformer, wang2022detr3d, yang2022bevformer, huang2022bevdet4d, Xiong_2023_CVPR} and LiDAR-based 3D detection~\cite{lang2019pointpillar, yin2021centerpoint, shi2021pvrcnn++}. 
Vision-based approaches~\cite{li2022bevstereo, wang2022sts, feng2022aedet, li2022bevdepth} exploit image information and frequently involve deep learning techniques to estimate depth. 
In contrast, LiDAR-based methods capitalize on precise geometric information from LiDAR sensors to achieve superior object detection accuracy. 
Our proposed vision-centric expert shares the same modality as vision-based detectors, while exhibiting superior performance compared to LiDAR-based detectors.

Long-term modeling has been employed to improve the performance of 3D object detection models~\cite{han2023recurrent, park2022solofusion}. 
SOLOFusion~\cite{park2022solofusion} utilizes long-term temporal modeling to achieve excellent performance. 
VideoBEV~\cite{han2023recurrent} maintains comparable performance with SOLOFusion while being more efficient, using long-term recurrent temporal modeling. 
However, previous research~\cite{li2022bevstereo, wang2023streampetr} has highlighted that long-term temporal fusion can lead to inadequate detection of dynamic objects. 
Although StreamPETR~\cite{wang2023streampetr} 
proposes the propagation transformer~\cite{vaswani2017attention, cmz2023cf, cmz2023diffrate, cmz2023smmix} to conduct object-centric temporal modeling, the improvement of dynamic object modeling remains relatively modest. 
Our proposed trajectory-based distillation module alleviates this limitation, enabling accurate detection of both static and dynamic objects by camera-based 3D object detection models that utilize long-term modeling.

\paragraph{Multi-modality Fusion.}
Multi-modality fusion techniques~\cite{vora2020pointpainting, bai2022transfusion, li2022uvtr, gao2023sparse} have been extensively investigated to enhance 3D object detection performance by integrating complementary information from different sensor modalities, such as cameras and LiDAR sensors. These methods~\cite{liang2022bevfusion, yang2022deepinteraction} typically require complex training strategies and the development of specialized fusion modules to effectively merge the distinct sources of information. In contrast, we propose a streamlined architecture that utilizes an image backbone for feature extraction, obviating the need for a LiDAR backbone. This efficient approach augments vision-based models by incorporating LiDAR information, maintaining homogeneity with vision-based models while preserving exceptional performance.

\paragraph{Knowledge Distillation.}

Knowledge distillation~\cite{romero2014fitnets, shu2021cwd} is a technique facilitating the transfer of knowledge from a larger, more complex model (expert) to a smaller, more efficient model (apprentice). This approach has been successfully applied in various domains, including image classification~\cite{deng2009imagenet}, natural language processing~\cite{vaswani2017attention}, and policy learning~\cite{wu2022trajectoryguided, jia2023thinktwice, jia2023driveadapter, chen2023end}. 
Therefore, recent distillation methods~\cite{guo2021liga, huang2022tigbev, li2022bevlgkd, huang2023geometricaware, klingner2023X3KD} build upon 3D object detection aims to transfer the accurate geometry knowledge from LiDAR to camera. MonoDistill~\cite{chong2022monodistill} projects the LiDAR points into the image plane to serve as the expert to transfer knowledge. BEVSimDet~\cite{zhao2023bevsimdet} simulates fusion-based methods to alleviate the domain gap between the two different modalities. BEVDistill~\cite{chen2022bevdistill} projects the LiDAR points and images into the BEV space to align the LiDAR feature and image feature. 
Due to the non-homogenous nature between the LiDAR and camera, transferring knowledge from LiDAR to images is challenging. Instead, our work constructs a vision-centric expert, which possesses a homogeneous modality with vision-based models. The vision-centric expert can leverage knowledge distillation to transfer the geometric perception capabilities to various vision-based models, hence enhancing performance accordingly.

\section{Method}
\label{headings}

In this section, we present our approach in detail. The overall architecture is presented in Sec.~\ref{overall_txt}. Our method involves two main components: (1) the vision-centric expert in Sec.~\ref{expert_txt}, and (2) the trajectory-based distillation and occupancy reconstruction modules as elaborated in Sec.~\ref{distillation_txt}.  The pipeline of our method is depicted in Fig.~\ref{pipelines}. 

\subsection{Overall Architecture}
\label{overall_txt}

In this paper, we construct a pair of harmonious expert and apprentice models. The expert and apprentice models adopt the consistent model architecture. The only difference is that the expert additionally leverages the accurate depth map generated from the point cloud, while the apprentice model predicts the depth map from the image. Although our expert model only uses an image backbone to encode high-level scene information, it is on par with state-of-the-art multi-model fusion methods that use several modality-specific backbones and complex interaction strategies. More importantly, we eliminate the domain gap between the multi-model expert and the camera-only apprentice model, which is deemed as one of the most challenging topics in the 
cross-modality distillation literature.

\begin{figure}
  \centering
  \includegraphics[width=.99\linewidth]{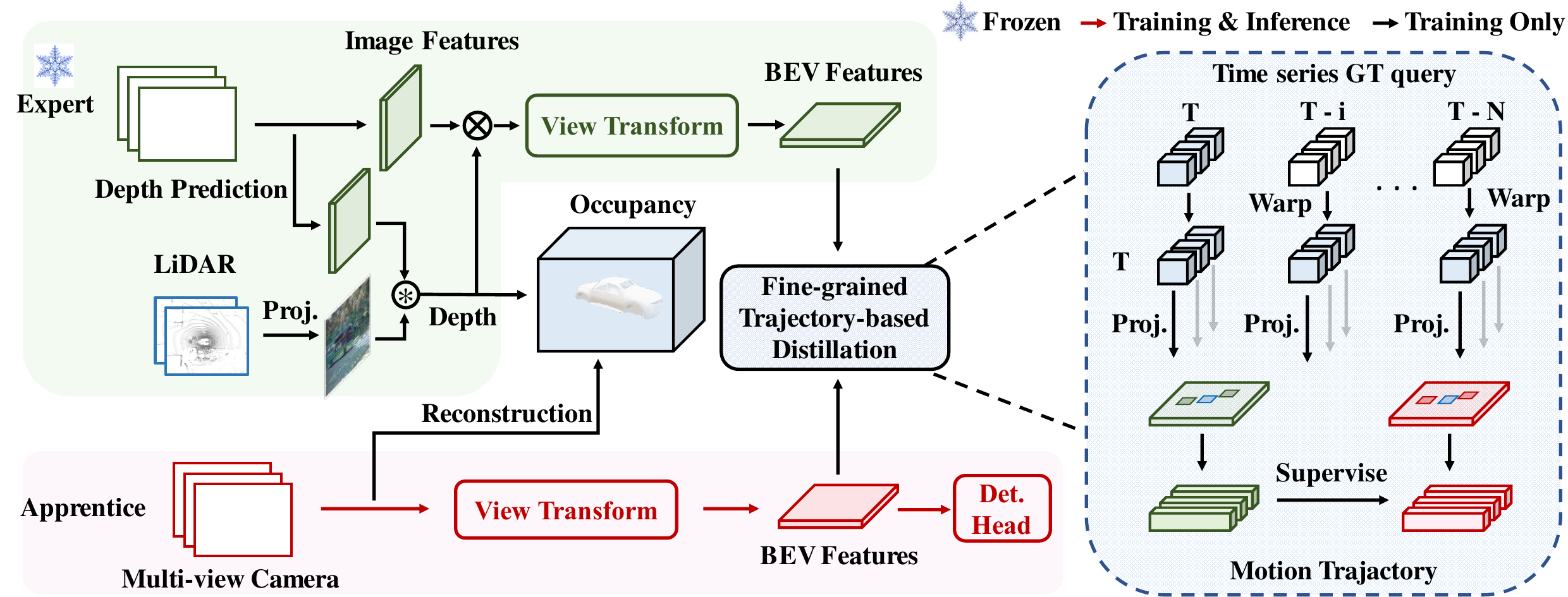}
  \caption{\textbf{Algorithm Overview.}
  Expert utilizes LiDAR data to enhance depth estimation accuracy before view transformation in BEV pipeline. 
  Apprentice represents a standard long-term vision-based detection model. 
  The occupancy is built using the depth information from the expert model, which serves as the supervision for the apprentice model. Motion trajectory is constructed by warping the time series GT query into the current timestamp. 
  Projecting the motion trajectory of each object into BEV space can rectify the misalignment of object motion. 
  With the knowledge transferred from the expert, the apprentice can deliver higher performance than before.}
  \label{pipelines}
\end{figure}

As illustrated in Fig.~\ref{pipelines}, we construct a distillation framework between the expert network and the apprentice network. The vision-centric expert fuses feature $\boldsymbol{I}^E$ extracted from the image backbone and the temporal depth map $\boldsymbol{D}$ projected from LiDAR points to create a unified BEV representation $\boldsymbol{F}^E$ used for 3D object detection. Therefore, although we adopt a cross-modality approach for 3D object detection, the resulting representation remains homogeneous with image modality features. 

After obtaining the pretrained vision-centric expert and corresponding apprentice network, we freeze the expert network and leverage its intermediate features as auxiliary supervision for the apprentice network.
Since current advanced vision-based detectors employ long-term temporal modeling to attain state-of-the-art performance, we utilize a standard long-term temporal vision-based detector based on BEVDepth as the apprentice model in our context.
The expert model also utilizes long-term temporal modeling to ensure consistency and achieve higher performance.

\subsection{The Generation of Expert Model}
\label{expert_txt}
We construct a vision-centric expert by integrating LiDAR information as accurate depth input into a vision-based model. Nevertheless, given the sparsity of LiDAR depth data, we rely on the predicted depth values obtained from images for pixels lacking LiDAR depth information. We also utilize future frames to further improve the performance of the expert model in the offline 3D detection setting. The vision-based detector serves as the primary model, and LiDAR information complements it by providing precise depth information. This approach eliminates the need for complex training strategies or custom-designed fusion modules, streamlining the fusion process.


For the expert model, we project the last sweep LiDAR frame with the current LiDAR frame onto images to obtain corresponding depth maps $\boldsymbol{D}$. 
Since the depth maps $\boldsymbol{D}$ generated from point clouds can not cover every pixel of the images, we also predict depth distribution for each pixel based on the image features. Then we will project the image features onto the BEV space to obtain BEV features $\boldsymbol{F}^E$ based on their depth. 
Furthermore, we transform the BEV features $\boldsymbol{F}^E_{T-N}$ from previous timestamps to the current BEV features $\boldsymbol{F}^E_{T}$, modeling the long-term relationship. $N$ denotes the time interval between the current frame and the history frame. 
The unified BEV features $\boldsymbol{F}^E$ are then combined to produce 3D object detection predictions $\boldsymbol{P}$. The model is trained using a multi-task loss function that considers both 3D detection loss $\mathcal{L}_{Det}$ and depth estimation loss $\mathcal{L}_{Depth}$.

\subsection{The Procedure of Distillation}
\label{distillation_txt}
In this section, we elucidate the methodology adopted to overcome the constraints inherent in long-term modeling for multi-camera 3D object detection. This is achieved through the incorporation of two innovative modules within the distillation process: the trajectory-based distillation module and the occupancy reconstruction module.

\paragraph{Trajectory-based Distillation.}
The trajectory-based distillation module aims to improve the detection of dynamic objects by focusing on the inconsistent portion of the objects' motion. 
For $i$-th historical frame  at timestamp $t_i$ that contains $K$ objects, extract the $j$-th ground truth object position $\boldsymbol{P}_i^j = (x_i^j, y_i^j, z_i^j, 1)^T$ in the ego coordinate system. Determine the actual ego motion matrix $\boldsymbol{M}_i$ between the current frame $t_0$ and each historical frame $t_i$. Apply the ego-motion transformation matrix $\boldsymbol{M}_i$ to the ground truth object positions $\boldsymbol{P}_i^j$ to obtain the transformed positions $\boldsymbol{P}_i^{j'}$ in the current frame's coordinate system:
\begin{equation}
  \boldsymbol{P}_i^{j'} = \boldsymbol{M}_i \times \boldsymbol{P}_i^j. 
\end{equation}
We amalgamate the transformed ground truth object positions, $\boldsymbol{P}_i^{j'}$, from all historical frames to construct the motion trajectories. The trajectory of $j$-th object can thus be represented as a sequence of object positions within the current frame  as $(\boldsymbol{P}_1^{j'}, \boldsymbol{P}_2^{j'}, \dots, \boldsymbol{P}_N^{j'})$,
where $N$ is the number of points on each motion trajectory.
Let $\boldsymbol{F}^E_{ij}$, $\boldsymbol{F}^A_{ij}$ represent the sampled features on identical point $\boldsymbol{P}_i^{j'}$ from the   expert BEV feature $\boldsymbol{F}^E$ and apprentice BEV feature $\boldsymbol{F}^A$, respectively. They are sampled via bilinear interpretation and then are normalized as :
\begin{equation}
    \boldsymbol{F}_{ij}^E = \text{norm}(\boldsymbol{F}^E(P_i^{j'})),\quad \boldsymbol{F}_{ij}^A = \text{norm}(\boldsymbol{F}^A(P_i^{j'})). 
\end{equation}
Finally, the trajectory-based distillation loss $\mathcal{L}_{TD}$  is computed between the normalized key sampled features:
\begin{equation}
   \mathcal{L}_{TD} =  \frac{1}{K\times N}\sum_{j=1}^K\sum_{i=1}^N L_2(\boldsymbol{{F}}_{ij}^A, {\boldsymbol{{F}}}_{ij}^E).
\end{equation}
Ultimately, by using the motion trajectory as queries, we conduct trajectory-based distillation on these representative positions. This approach enables the expert to rectify the motion misalignment in the apprentice.

\paragraph{Occupancy Reconstruction.}
The expert model demonstrates outstanding performance in 3D object detection, resulting in a more precise 3D geometric representation of the objects.
We utilize depth estimation modules to predict the depth, denoted as $\boldsymbol{D}(u,v)$, for each image pixel $(u, v)$. Subsequently, the depth map $\boldsymbol{D}$ is back-projected into a 3D point cloud, each image pixel $(u,v)$ is transformed into a 3D coordinate (x, y, z). Then the scores from different pixels in the multi-camera within the same occupancy region, are accumulated to make decisions about the presence of an object in that region. In this way, we can generate the occupancy $\boldsymbol{O}_e$ and $\boldsymbol{O}_a$ for expert and apprentice model, respectively.

The occupancy reconstruction module improves the model's capability to discern the 3D geometric properties of objects. The grid-based supervisory signals effectively direct the model to enhance its prediction accuracy for object depths. Different voxels within the occupancy structure aggregate the fused depth distribution from various perspective views, thereby enhancing robustness against depth errors.
Drawing inspiration from CenterPoint~\cite{zhou2019center}, we simply extend the Gaussian distribution applied to each target into the 3D space for more focused 3D object modeling:
\begin{equation}
    \boldsymbol{G}_{xyz} = \text{exp}\bigg [-\frac{(x-\tilde{p}_x)^2 + (y-\tilde{p}_y)^2 + (z-\tilde{p}_z)^2}{2\sigma_p^2} \bigg ], 
\end{equation}
where $(\tilde{p}_x, \tilde{p}_y, \tilde{p}_z)$ represents the center of the 3D object, while $\sigma_p$ denotes the standard deviation of each object's size. The model utilizes the expert model's occupancy $O_{e}$ as supplementary supervision by adopting a straightforward $\mathcal{L}_1$ regularization loss for the occupancy status $O_{a}$ of the apprentice model, which optimizes depth prediction capabilities for both static and dynamic objects. The occupancy reconstruction loss can be formulated as
\begin{equation}
\mathcal{L}_{OR} = \mathcal{L}_1(\boldsymbol{G}_{xyz} \cdot \boldsymbol{O}_e, \boldsymbol{G}_{xyz}\cdot \boldsymbol{O}_a ).
\end{equation}

\paragraph{Training Loss.} During the distillation  phase, the joint training loss $\mathcal{L}_{Total}$ is formulated as 
\begin{equation}
\mathcal{L}_{Total} = \mathcal{L}_{A}  + \lambda_1 \cdot \mathcal{L}_{TD} + \lambda_2 \cdot \mathcal{L}_{OR}, 
\end{equation}
where $\mathcal{L}_{A}$ is the perceptual loss of the apprentice model. Besides, $\lambda_1$ and $\lambda_2$ represent hyperparameters employed to effectively balance the scales of the respective loss functions. The utilization of trajectory-based distillation loss $\mathcal{L}_{TD}$ and occupancy reconstruction loss  
$\mathcal{L}_{OR}$ collectively facilitates the transfer of semantic and geometry knowledge from the expert model to the apprentice model.


\begin{table}[tb]
  \caption{\textbf{Comparison among the camera-only methods on the nuScenes \textit{val} set.} $^*$ denotes the long-term baseline implemented by us based on BEVDet4D-Depth~\cite{huang2022bevdet4d}. $^\dagger$ depicts that the size of BEV feature is 256$\times$256. VCD-A surpasses previous SOTA by 2 points in NDS and achieves SOTA under the same setting.}
  \label{camera-val}
  \centering
  \resizebox{\linewidth}{22.5mm}{
  \begin{tabular}{{l|c|c|c|c|c|ccccc}}
    \toprule
    Methods  & Backbone  & Image Size & Frames & mAP$\uparrow$ & NDS$\uparrow$ & mATE$\downarrow$ & mASE$\downarrow$ & mAOE$\downarrow$ & mAVE$\downarrow$ & mAAE$\downarrow$ \\
    \midrule
    BEVDet~\cite{huang2021bevdet} & ResNet-50 & 256 $\times$ 704 & 1 & 0.298 & 0.379 & 0.725 & 0.279 & 0.589 & 0.860 & 0.245   \\
    PETR~\cite{liu2022petr} & ResNet-50 & 384 $\times$ 1056 & 1 & 0.313 & 0.381 & 0.768 & 0.278 & 0.564 & 0.923 & 0.225   \\
    BEVDet4D~\cite{huang2022bevdet4d} & ResNet-50 & 256 $\times$ 704 & 2 & 0.322 & 0.457 & 0.703 & 0.278 & 0.495 & 0.354 & 0.206 \\
    BEVDepth~\cite{li2022bevdepth} & ResNet-50 & 256 $\times$ 704 & 2 & 0.351 & 0.475 & 0.639 & 0.267 & 0.479 & 0.428 & 0.198 \\
    BEVStereo~\cite{li2022bevstereo} & ResNet-50 & 256 $\times$ 704 & 2 & 0.372 & 0.500 & 0.598 & 0.270 & 0.438 & 0.367 & 0.190 \\
    STS~\cite{wang2022sts} & ResNet-50 & 256 $\times$ 704 & 2 & 0.377 & 0.489 & 0.601 & 0.275 & 0.450 & 0.446 & 0.212  \\
    VideoBEV~\cite{han2023videobev} & ResNet-50    & 256 $\times$ 704 & 8 & 0.422 & 0.535 & 0.564 & 0.276 & 0.440 & 0.286 & 0.198 \\
    SOLOFusion~\cite{park2022solofusion} & ResNet-50 & 256 $\times$ 704 & 16+1 & 0.427 & 0.534 & 0.567 & 0.274 & 0.411 & \textbf{0.252} & \textbf{0.188} \\
    StreamPETR~\cite{wang2023streampetr} & ResNet-50 & 256 $\times$ 704 & 8 & 0.432 & 0.540 & 0.581 & 0.272 & 0.413 &  0.295 & 0.195 \\
    \midrule
    Baseline$^*$& ResNet-50 & 256 $\times$ 704 & 8+1 & 0.401 & 0.515 & 0.595 & 0.279 & 0.489 & 0.291 & 0.198  \\
    \gray{VCD-A} & \gray ResNet-50 & \gray 256 $\times$ 704 & \gray 8+1 & \gray0.426 & \gray0.540 & \gray0.547 & \gray0.271 & \gray0.433 & \gray0.268 & \gray0.207 \\
    Baseline$^{*\dagger}$ & ResNet-50 & 256 $\times$ 704 & 8+1 & 0.418 & 0.542 & 0.522 & 0.267 & 0.428 & 0.262 & \textbf{0.188}  \\
    \gray VCD-A$^\dagger$ & \gray ResNet-50 & \gray 256 $\times$ 704 & \gray8+1 & \gray\textbf{0.446} & \gray\textbf{0.566} & \gray\textbf{0.497} & \gray\textbf{0.260} & \gray\textbf{0.350} & \gray0.257 & \gray0.203 \\
    \bottomrule
  \end{tabular}
  }
\end{table}

\begin{table}[tb]
  \caption{\textbf{Comparison among the camera-only methods on the nuScenes \textit{test} set.} Methods marked with $^*$ denote long-term baseline implemented by us, based on BEVDet4D-Depth~\cite{huang2022bevdet4d}. $\dagger$ depicts test time augmentation adopted during the inference phase. VCD-A achieves SOTA under critical metrics and surpasses its baseline by 2 points in NDS.}
  \label{camera-test}
  \centering
  \resizebox{\linewidth}{24.5mm}{
  \begin{tabular}{{l|c|c|c|c|ccccc}}
    \toprule
    Methods  & Backbone  & Image Size & mAP$\uparrow$ & NDS$\uparrow$ & mATE$\downarrow$ & mASE$\downarrow$ & mAOE$\downarrow$ & mAVE$\downarrow$ & mAAE$\downarrow$ \\
    \midrule
    FCOS3D$\dagger$~\cite{wang2021fcos3d} & R101-DCN & 900 $\times$ 1600  & 0.358 & 0.428 & 0.690 & 0.249 & 0.452 & 1.434 & 0.124   \\
    DETR3D$\dagger$~\cite{wang2022detr3d} & V2-99 & 900 $\times$ 1600 & 0.412 & 0.479 & 0.641 & 0.255 & 0.394 & 0.845 & 0.133   \\
    UVTR~\cite{li2022uvtr} & V2-99 & 900 $\times$ 1600 & 0.472 & 0.551 & 0.577 & 0.253 & 0.391 & 0.508 & 0.123 \\
    BEVDet4D$\dagger$~\cite{huang2022bevdet4d} & Swin-B~\cite{liu2021swin} & 900 $\times$ 1600 & 0.451 & 0.569 & 0.511 & \textbf{0.241} & 0.386 & 0.301 & 0.121 \\
    BEVFormer~\cite{li2022bevformer} & V2-99 & 900 $\times$ 1600 & 0.481 & 0.569 & 0.582 & 0.256 & 0.375 & 0.378 & 0.126 \\
    PolarFormer~\cite{jiang2022polarformer} & V2-99 & 900 $\times$ 1600 & 0.493 & 0.572 & 0.556 & 0.256 & 0.364 & 0.439 & 0.127  \\
    BEVDistill~\cite{chen2022bevdistill} & ConvNeXt-B & 900 $\times$ 1600 & 0.496 & 0.594 & 0.475 & 0.249 & 0.378 & 0.313 & 0.125  \\
    PETRv2~\cite{liu2022petrv2}& RevCol~\cite{cai2022reversible} & 640 $\times$ 1600 & 0.512 & 0.592 & 0.547 & 0.242 & 0.360 & 0.367 & 0.126  \\
    BEVDepth~\cite{li2022bevdepth} & ConvNeXt-B & 640 $\times$ 1600 & 0.520 & 0.609 & 0.445 & 0.243 & 0.352 & 0.347 & 0.127 \\
    AeDet$\dagger$~\cite{feng2022aedet} & ConvNeXt-B & 640 $\times$ 1600 & 0.531 & 0.620 & 0.439 & 0.247 & 0.344 & 0.292 & 0.130 \\
    SOLOFusion~\cite{park2022solofusion} & ConvNeXt-B & 640 $\times$ 1600 & 0.540 & 0.619 & 0.453 & 0.257 & 0.376 & 0.276 & 0.148 \\
    StreamPETR~\cite{wang2023streampetr} & ConvNeXt-B & 640 $\times$ 1600 & \textbf{0.550} & \textbf{0.631} & 0.493 & \textbf{0.241} & \textbf{0.343} & \textbf{0.243} & 0.123 \\
    \midrule
    Baseline$^*$ &  ConvNeXt-B & 640 $\times$ 1600 & 0.522 & 0.610 & 0.457 & 0.253 & 0.391 & 0.271 & 0.142 \\
    \gray VCD-A & \gray ConvNeXt-B & \gray 640 $\times$ 1600 & \gray0.548 & \gray\textbf{0.631} & \gray\textbf{0.436} & \gray0.244 & \gray\textbf{0.343} & \gray0.290 & \gray\textbf{0.120} \\
    \bottomrule
  \end{tabular}
  }
  \vspace{-3mm}
  \end{table}

\section{Experiments}
\label{others}

In this section, we outline the experimental setup and assess our proposed VCD-E model, as well as the newly introduced trajectory-based distillation and occupancy reconstruction modules, using the nuScenes dataset~\cite{caesar2020nuscenes}. This includes a presentation of the evaluation metrics, baseline models, ablation studies, and a comparative analysis of our approach with the current state-of-the-art methods.

\subsection{Main Results}

\paragraph{Camera-only 3D detection.}
In order to evaluate the effectiveness of our proposed Revenge model and the trajectory-based distillation modules, we have conducted rigorous experiments on the nuScenes validation and test sets. As presented in Tab.~\ref{camera-val}, the performance of VCD-A surpasses other cutting-edge methods, achieving a record of 44.6\% and 56.6\% on the nuScenes benchmark. This provides robust evidence of the effectiveness of our approach. Utilizing an image resolution of 256$\!\times\!$704 and a ResNet-50 backbone, VCD outperforms the state-of-the-art method from \cite{park2022solofusion} by improving mAP by 1.9\% and NDS by 3.2\%. 
Additionally, VCD-A exhibits an improvement over our baseline by 2.8\% mAP and 2.4\% NDS, thereby indicating that our methods can considerably advance state-of-the-art results.
To further ascertain the generalizability of our methods, we conducted experiments on the nuScenes test set, as shown in Tab.~\ref{camera-test}. With the adoption of the ConvNext-B backbone~\cite{liu2022convnet}, VCD achieved 54.8\% mAP and 63.1\% NDS, outperforming the state-of-the-art detector, SOLOFusion~\cite{park2022solofusion}, by 0.8\% mAP and 1.2\% NDS. Furthermore, our proposed approach led to a 2.6\% increase in mAP and a 2.1\% improvement in NDS compared to our baseline, thereby demonstrating the efficacy of our method for large-scale vision-based models.

\paragraph{Multi-modality Fusion.}
We first compare our proposed vision-centric expert with other state-of-the-art multi-modality fusion methods. VCD-E does not require any complex fusion strategies or three-stage training schedule, yet it achieves 67.7\% mAP and a 71.1\% NDS as shown in Tab.~\ref{fusion-val}. Despite employing only a single image backbone, the model's performance is comparable to state-of-the-art methods such as BEVFusion~\cite{liang2022bevfusion}, and it even surpasses UVTR~\cite{li2022uvtr} by 2.3\% mAP and 0.9\% NDS, which utilizes two backbones, by a considerable margin. Owing to its homogeneous nature and compatibility with camera-only models, the VCD-E can serve as a powerful expert for knowledge transfer, facilitating performance improvements for small to large vision detectors.

\begin{table}[tb]
  \caption{\textbf{Comparison among the multi-modality methods on the nuScenes \textit{val} set.} Our proposed VCD-E only adopts image backbone while achieving comparable performance with state-of-the-art multi-modal methods who predominately rely on the LiDAR backbone.}
  \label{fusion-val}
  \centering
  \resizebox{\linewidth}{10.8mm}{
  \begin{tabular}{{l|c|c|c|c|ccccc}}
    \toprule
    Methods  & Venue  & Backbone & mAP$\uparrow$ & NDS$\uparrow$ & mATE$\downarrow$ & mASE$\downarrow$ & mAOE$\downarrow$ & mAVE$\downarrow$ & mAAE$\downarrow$ \\
    \midrule
    BEVFusion~\cite{liang2022bevfusion} & NeurIPS~2022
    & LiDAR \& Image & 0.642 & 0.680 & - & - & - &  - & - \\
    FUTR3D~\cite{chen2022futr3d} & Arxiv~2022
    & LiDAR \& Image & 0.645 & 0.683 & - & - & - & - & -   \\
    UVTR~\cite{li2022uvtr} & NeurIPS~2022
    & LiDAR \& Image & 0.654 & 0.702 & 0.332 & 0.258 & \textbf{0.268} & 0.212 & \textbf{0.177}   \\
    CMT~\cite{yan2023cmt} & Arxiv~2023
    & LiDAR \& Image & \textbf{0.679} & 0.708 & - & - & - & - & -   \\
     \gray VCD-E & \gray - & \gray Image & \gray0.677 & \gray\textbf{0.711} & \gray\textbf{0.308} & \gray\textbf{0.254} & \gray0.317 & \gray\textbf{0.189} & \gray0.201 \\
    \bottomrule
  \end{tabular}
  }
  \vspace{-3mm}
\end{table}

\begin{table}[tb]
  \caption{\textbf{Ablation study of the proposed distillation framework on different temporal lengths.}
  The design works with different length of time windows and the gain grows with the temporal length.
  }
  \label{temporal}
  \centering
  \resizebox{0.75\linewidth}{19.mm}{
  \begin{tabular}{{c|c|l|l|ccc}}
  \toprule
  {Temporal Length}   & {Distill} & mAP (\%)$\uparrow$ & NDS (\%) $\uparrow$ & mATE$\downarrow$ & mAOE$\downarrow$ & mAVE$\downarrow$\\ 
  \midrule
  \multirow{ 2}{*}{1}  &\xmark  & 26.6 & 37.9 & 0.815 & 0.645 & 0.556\\ 
   & \cmark  & 30.1 \small{(+3.5)} & 41.5\small{(+3.6)} & {0.732} & {0.629} & {0.476} \\ 
  \midrule
  \multirow{ 2}{*}{2}  & \xmark & 26.9 & 38.4 & 0.804 & 0.706 & 0.461\\ 
   & \cmark  & 31.3 \small{(+4.4)} & 43.2 \small{(+4.8)} & {0.717} & {0.615} & {0.403} \\ 
  \midrule
  \multirow{ 2}{*}{4}  & \xmark & 28.4 & 39.8 & 0.748 & 0.739 & 0.432 \\
   & \cmark & 33.0 \small{(+4.6)} & 44.1 \small{(+4.3)} & {0.707} & {0.632} & {0.389} \\
  \midrule
  \multirow{ 2}{*}{8} &  \xmark& 29.7 & 40.9 & 0.762 & 0.714 & 0.415\\
   & \cmark & 35.4 \small{(+5.7)} & 45.9 \small{(+5.0)}  & {0.690} & {0.625} & {0.370}  \\
  \bottomrule
  \end{tabular}
  }
\vspace{-3mm}  
\end{table}

\subsection{Ablation Study}
To verify the effectiveness and necessity of each component, we conduct various ablation experiments on the nuScenes validation set. 

\paragraph{Effectiveness of the General Framework.}
We first validate the effectiveness of the general framework for various temporal lengths. In Tab.~\ref{temporal}, we examine different timestamps for temporal modeling and find that our method significantly outperforms the baseline. By employing the general framework, we achieve 5.0\% NDS and 5.7\% mAP performance gains with 8 temporal modeling instances. For other temporal modeling scenarios, our method continues to exhibit substantial performance improvements, ranging from 3.6\% to 5.0\% NDS. As the duration of temporal fusion extends, the advantages of the model become increasingly pronounced, suggesting that this framework exhibits higher compatibility with extended sequences. This indicates the effectiveness of our proposed trajectory-based distillation module, which alleviates the issue of motion misalignment.

\paragraph{Effectiveness of VCD-E.}
In this study, it is crucial to verify the effectiveness of our proposed vision-centric experts. We select various expert models for a fair comparison, including LiDAR-based expert Centerpoint~\cite{yin2021centerpoint}, vision-based expert BEVDepth~\cite{li2022bevdepth}, and fusion-based expert Transfusion, based on our proposed distillation strategy. In Tab.~\ref{experts}, we observe that homogeneous expert models significantly influence the success of knowledge transfer. The camera-based expert BEVDepth demonstrates superior performance gains compared to the other two heterogeneous expert models. Our advanced model, VCD-E, operates within the same modality as vision-based models, and it is equipped with precise geometric information. Consequently, the distillation effect outperforms the Transfusion~\cite{bai2022transfusion} expert by a significant margin, achieving a 6.2\% increase in mAP. Moreover, it surpasses the BEVDepth expert by an additional 1.7\% in NDS. This demonstrates the superior performance and effectiveness of our model.

\paragraph{Comparison to SOTA.}
To further demonstrate the effectiveness of our proposed distillation strategy, we compare our method with other state-of-the-art methods. To ensure a fair comparison, we conduct all experiments based on our vision-centric expert VCD-E. We find that our method consistently outperforms other distillation methods by a significant margin. In 2D detection, FitNet~\cite{romero2014fitnets} and CWD~\cite{shu2021cwd} are two classic distillation methods, we adapt them for 3D object detection to facilitate a fair comparison. BEVDistill~\cite{chen2022bevdistill} represents a state-of-the-art distillation method for multi-view 3D object detection, and we compare our results with this approach as well. As shown in Tab.~\ref{methods}, our method achieves the best results when compared to these state-of-the-art distillation strategies, demonstrating the effectiveness of our approach. Our method surpasses BEVDistill by 2\% NDS and 3.8\% mAP.

\begin{table}[t]
 \begin{minipage}{0.5\linewidth}
 \centering
 \caption{
 \textbf{The performance gains of the apprentice} which benefit from different experts. CM denotes cross-modal and UM represents Uni-modal. It indicates the success of uni-modal distillation remains.
 }
 \label{experts}
 \setlength{\tabcolsep}{4pt}
 
\begin{tabular}{l|c|c|c}
  \toprule
  {Expert}   & {Paradigm} & mAP & NDS \\ 
  \midrule
  - & - & 0.297 & 0.409 \\
  CenterPoint~\cite{yin2021centerpoint}  & CM  & 0.281 & 0.420 \\ 
  Transfusion~\cite{bai2022transfusion}  & CM & 0.292 & 0.435 \\
  BEVDepth~\cite{li2022bevdepth}  & UM & 0.341 & 0.442 \\ 
  VCD-E & UM & \textbf{0.354} & \textbf{0.459}  \\
  \bottomrule
 \end{tabular}
 \end{minipage}
 \hfill
 \begin{minipage}{0.45\linewidth}
\centering
 \caption{
 \textbf{Effect of different distillation methods.} All models are trained with VCD-E as expert. Our proposal significantly surpasses previous SOTA methods.
 }
 \label{methods}
 \setlength{\tabcolsep}{6pt}

\begin{tabular}{l|c|c}
  \toprule
  {Methods}  & mAP & NDS \\ 
  \midrule
  Baseline~\cite{li2022bevdepth} & 0.297 & 0.409 \\
  FitNet~\cite{romero2014fitnets} & 0.318	& 0.421 \\ 
  CWD~\cite{shu2021cwd} & 0.311 & 0.412 \\ 
  BEVDistill~\cite{chen2022bevdistill} & 0.316 & 0.439 \\
  VCD-A & \textbf{0.354} & \textbf{0.459} \\
  \bottomrule
 \end{tabular}
 \end{minipage}

\end{table}

\begin{table}[t]
\begin{minipage}{0.5\linewidth}
\centering
 \caption{
 \textbf{Gains of different image backbone} on multi-modal models~\cite{liang2022bevfusion}. The stronger backbone still demonstrates better performance in this case.
 }
 \label{expert}
 \setlength{\tabcolsep}{6pt}

\begin{tabular}{l|c|c|c}
  \toprule
  {Methods}   & {Backbone} & mAP & NDS \\ 
  \midrule
  BEVFusion  & ResNet-50 & 0.598 & 0.662 \\ 
  BEVFusion  & ConvNext-B  & 0.597 & 0.665 \\ 
  VCD-E  & ResNet-50 & 0.611 & 0.656 \\
  VCD-E & ConvNext-B & \textbf{0.664} & \textbf{0.693}  \\
  \bottomrule
 \end{tabular}
 \end{minipage}
 \hfill
 \begin{minipage}{0.45\linewidth}
\centering
 \caption{
 \textbf{Gains of different depth fusion strategy.}
 The proposed fusion depth is optimal among different methods.
 }
 \label{depth}
 \setlength{\tabcolsep}{6pt}

\begin{tabular}{l|c|c}
  \toprule
  {Methods}   &  mAP & NDS \\ 
  \midrule
  Predicted depth  &  0.495 & 0.585 \\ 
  LiDAR depth  & 0.638  & 0.687 \\ 
  Weighted depth & 0.644 & \textbf{0.690} \\
  Fusion depth  &  \textbf{0.646} & \textbf{0.690} \\
  \bottomrule
 \end{tabular}
 \end{minipage}

\end{table}

\vspace{-3mm}
\paragraph{Gains of the Image Backbone.}

To verify that VCD-E benefits from contemporary image backbones, we conducted experiments outlined in Tab.~\ref{expert}. We selected ResNet-50 and ConvNext-B as two distinct modern image backbones for BEVFusion and VCD-E. Our findings indicate that VCD-E achieves substantial improvements of 5.3\% mAP and 3.7\% NDS when using ConvNext-B compared to ResNet-50, demonstrating that VCD-E can indeed benefit from modern image backbones. However, existing fusion-based methods, such as BEVFusion, which primarily rely on the capabilities of LiDAR backbones, show limited gains from employing modern image backbones.

\paragraph{Fusion Strategy of Expert.}
To determine the optimal fusion strategy for LiDAR points and images, we propose four distinct depth fusion approaches, including (1) Predicted depth that directly uses predicted depth based on image features. (2) LiDAR depth that utilizes the projected depth from LiDAR points. (3) Fusion depth that employs the projected depth when corresponding LiDAR points exist for each pixel, otherwise using the predicted scores based on image features, and (4) Weighted depth. Applying the weighted average depth of the predicted depth and LiDAR depth. As demonstrated in Tab.~\ref{depth}, fusion depth yields the best results which indicates that using corresponding predicted scores where LiDAR depth is unavailable is advantageous.

\paragraph{The Effectiveness of Trajectory-based Distillation}
The results presented in Table~\ref{trajectory_length} indicate that as the trajectory length increases, the benefits derived from the distillation process become more pronounced. The temporal fusion length for this experiment is set at eight. The first row denotes the baseline model which does not use the distillation method. The second row depicts that the VCD-A model directly conducts distillation under the full BEV feature without the Trajectory-based distillation module. When the trajectory length is set to 1, we only distill ground truth locations in the current frame. However, when the trajectory length exceeds five, there is a noticeable decrease in accuracy. We hypothesize that this decrease may be attributed to the model's distracted attention towards distant motions. The density of traffic can lead to distant motion locations being occupied by other objects, which may not necessarily require additional trajectory supervision. This suggests that the application of excessive trajectory supervision in such scenarios could be unnecessary and inefficient.

\begin{table}[t]
  \caption{\textbf{The performance gains of different trajectory length} for trajectory-based distillation. As the trajectory length increases, the benefits derived from the distillation process become more pronounced.
  }
  \label{trajectory_length}
  \centering
  \begin{tabular}{{c|c|c|c}}
  \toprule
  Trajectory Length   & Distill &  mAP (\%) & NDS (\%) \\ 
  \midrule
   -  & \xmark   & 29.7 & 40.9 \\ 
   0 & \cmark  & 31.8 & 42.1 \\ 
   1 & \cmark  & 33.1 & 44.5 \\ 
   3 & \cmark  & 34.6  & 45.6 \\ 
   5 & \cmark  & \textbf{35.4}  & \textbf{45.9} \\ 
   9 & \cmark  & 33.9  & 44.7 \\ 
  \bottomrule
  \end{tabular}
\end{table}

\paragraph{Ablation Study of Each Component.}

In Tab.~\ref{whistles}, we conduct an ablation study on the components employed in VCD to verify their contributions to the final result. These components include (1) Long-term temporal fusion, which allows our model to benefit from historical information, thereby enhancing its performance. (2) Trajectory-based distillation, which transfers knowledge from VCD-E to VCD-A in areas with motion misalignment to mitigate this issue. 
(3) Occupancy reconstruction, which serves as dense depth supervision in the perspective view, improving the performance of VCD-A. Long-term temporal modeling enables our model to reference prior information from history, assisting in object velocity estimation and resulting in a significantly lower mATE. We also evaluate the impact of the trajectory-based distillation module when all other components are incorporated. As shown in Tab.~\ref{whistles}, trajectory-based distillation increases NDS by 3.3\% and mAP by 3.3\%, contributing to the majority of the improvement.

\begin{table}
  \caption{\textbf{Ablation study of important components in VCD-A.} It shows that the motion module contributes most to the performance improvement, indicating the potential of future research here.}
  \label{whistles}
  \centering
  \resizebox{0.93\linewidth}{14.8mm}{
  \begin{tabular}{{l|ccc|c|c|cc}}
    \toprule
    Methods   & Long & Occ & Motion & mAP$\uparrow$ & NDS$\uparrow$ & mATE$\downarrow$ &  mAVE$\downarrow$  \\
    \midrule
    Baseline &  &  &  & 0.271 & 0.319 & 0.767 & 1.065 \\
    Longer Temporal Fusion & \cmark &  &  & 0.297 & 0.409 & 0.762 &  0.415   \\
    Trajectory-based Distillation &  &  & \cmark & 0.283 & 0.356 & 0.772 & 0.870 \\
    All but Longer Temporal &  & \cmark & \cmark & 0.290 & 0.367 & 0.724 & 0.834 \\
    Occupancy Reconstruction  & \cmark & \cmark &  & 0.308 & 0.416 & 0.734 & \textbf{0.379}  \\
   VCD & \cmark & \cmark & \cmark & \textbf{0.341} & \textbf{0.449} & \textbf{0.715}  & 0.389 \\
    \bottomrule
  \end{tabular}
  }
  \vspace{-3mm}
\end{table}


\section{Conclusion}

In this paper, we presented a novel vision-centric model as the expert, which leverages the strengths of both modalities, our proposed model achieves exceptional performance, rivaling that of state-of-the-art multimodal fusion models, while also mitigating domain gap issues, making it suitable for distilling vision-based models.
Furthermore, we introduced two innovative modules, the trajectory-based distillation and occupancy reconstruction modules. These modules enhance the geometric perception capabilities of multi-camera 3D object detection models and improve the detection of both static and dynamic objects in the scene.
Our experiments demonstrate the effectiveness of our proposed methods on the widely-used nuScenes 3D object detection benchmark. 

\paragraph{Limitation.}
In this work, we have not delved into more details of the vision-centric expert, which may hold significant potential for improvement. Additionally, we have not explored further applications, such as automatic dimension estimation based on VCD-E.

\section*{Acknowledgements}
This work was supported by National Key R\&D Program of China (2022ZD0160104) and NSFC (62206172). We would like to thank anonymous reviewers for active discussions.

\bibliographystyle{plain}
\bibliography{bibliography_long, bibliography}

\begin{thebibliography}{10}

\bibitem{ba2016layernorm}
Jimmy~Lei Ba, Jamie~Ryan Kiros, and Geoffrey~E Hinton.
\newblock Layer normalization.
\newblock {\em arXiv preprint arXiv:1607.06450}, 2016.

\bibitem{bai2022transfusion}
Xuyang Bai, Zeyu Hu, Xinge Zhu, Qingqiu Huang, Yilun Chen, Hongbo Fu, and
  Chiew-Lan Tai.
\newblock {TransFusion}: Robust lidar-camera fusion for 3d object detection
  with transformers.
\newblock In {\em CVPR}, 2022.

\bibitem{caesar2020nuscenes}
Holger Caesar, Varun Bankiti, Alex~H Lang, Sourabh Vora, Venice~Erin Liong,
  Qiang Xu, Anush Krishnan, Yu~Pan, Giancarlo Baldan, and Oscar Beijbom.
\newblock {nuScenes}: A multimodal dataset for autonomous driving.
\newblock In {\em CVPR}, 2020.

\bibitem{cai2022reversible}
Yuxuan Cai, Yizhuang Zhou, Qi~Han, Jianjian Sun, Xiangwen Kong, Jun Li, and
  Xiangyu Zhang.
\newblock Reversible column networks.
\newblock In {\em ICLR}, 2023.

\bibitem{chen2022persformer}
Li~Chen, Chonghao Sima, Yang Li, Zehan Zheng, Jiajie Xu, Xiangwei Geng,
  Hongyang Li, Conghui He, Jianping Shi, Yu~Qiao, et~al.
\newblock Persformer: 3d lane detection via perspective transformer and the
  openlane benchmark.
\newblock In {\em ECCV}, 2022.

\bibitem{chen2023end}
Li~Chen, Penghao Wu, Kashyap Chitta, Bernhard Jaeger, Andreas Geiger, and
  Hongyang Li.
\newblock End-to-end autonomous driving: Challenges and frontiers.
\newblock {\em arXiv preprint arXiv:2306.16927}, 2023.

\bibitem{cmz2023cf}
Mengzhao Chen, Mingbao Lin, Ke~Li, Yunhang Shen, Yongjian Wu, Fei Chao, and
  Rongrong Ji.
\newblock Cf-vit: A general coarse-to-fine method for vision transformer.
\newblock In {\em AAAI}, 2023.

\bibitem{cmz2023smmix}
Mengzhao Chen, Mingbao Lin, Zhihang Lin, Yuxin Zhang, Fei Chao, and Rongrong
  Ji.
\newblock Smmix: Self-motivated image mixing for vision transformers.
\newblock In {\em ICCV}, 2023.

\bibitem{cmz2023diffrate}
Mengzhao Chen, Wenqi Shao, Peng Xu, Mingbao Lin, Kaipeng Zhang, Fei Chao,
  Rongrong Ji, Yu~Qiao, and Ping Luo.
\newblock Diffrate: Differentiable compression rate for efficient vision
  transformers.
\newblock {\em arXiv preprint arXiv:2305.17997}, 2023.

\bibitem{chen2022futr3d}
Xuanyao Chen, Tianyuan Zhang, Yue Wang, Yilun Wang, and Hang Zhao.
\newblock Futr3d: A unified sensor fusion framework for 3d detection.
\newblock {\em arXiv preprint arXiv:2203.10642}, 2022.

\bibitem{chen2022bevdistill}
Zehui Chen, Zhenyu Li, Shiquan Zhang, Liangji Fang, Qinhong Jiang, and Feng
  Zhao.
\newblock Bevdistill: Cross-modal bev distillation for multi-view 3d object
  detection.
\newblock {\em arXiv preprint arXiv:2211.09386}, 2022.

\bibitem{chong2022monodistill}
Zhiyu Chong, Xinzhu Ma, Hong Zhang, Yuxin Yue, Haojie Li, Zhihui Wang, and
  Wanli Ouyang.
\newblock {MonoDistill}: Learning spatial features for monocular 3d object
  detection.
\newblock {\em arXiv preprint arXiv:2201.10830}, 2022.

\bibitem{mmdet3d2020}
MMDetection3D Contributors.
\newblock {MMDetection3D: OpenMMLab} next-generation platform for general {3D}
  object detection.
\newblock \url{https://github.com/open-mmlab/mmdetection3d}, 2020.

\bibitem{deng2009imagenet}
Jia Deng, Wei Dong, Richard Socher, Li-Jia Li, Kai Li, and Li~Fei-Fei.
\newblock {ImageNet}: A large-scale hierarchical image database.
\newblock In {\em CVPR}, 2009.

\bibitem{feng2022aedet}
Chengjian Feng, Zequn Jie, Yujie Zhong, Xiangxiang Chu, and Lin Ma.
\newblock Aedet: Azimuth-invariant multi-view 3d object detection.
\newblock {\em arXiv preprint arXiv:2211.12501}, 2022.

\bibitem{gao2023sparse}
Yulu Gao, Chonghao Sima, Shaoshuai Shi, Shangzhe Di, Si~Liu, and Hongyang Li.
\newblock Sparse dense fusion for 3d object detection.
\newblock In {\em IROS}, 2023.

\bibitem{guo2021liga}
Xiaoyang Guo, Shaoshuai Shi, Xiaogang Wang, and Hongsheng Li.
\newblock {LIGA-Stereo}: Learning lidar geometry aware representations for
  stereo-based 3d detector.
\newblock In {\em ICCV}, 2021.

\bibitem{han2023recurrent}
Chunrui Han, Jianjian Sun, Zheng Ge, Jinrong Yang, Runpei Dong, Hongyu Zhou,
  Weixin Mao, Yuang Peng, and Xiangyu Zhang.
\newblock Exploring recurrent long-term temporal fusion for multi-view 3d
  perception.
\newblock {\em arXiv preprint arXiv:2303.05970}, 2023.

\bibitem{han2023videobev}
Chunrui Han, Jianjian Sun, Zheng Ge, Jinrong Yang, Runpei Dong, Hongyu Zhou,
  Weixin Mao, Yuang Peng, and Xiangyu Zhang.
\newblock Exploring recurrent long-term temporal fusion for multi-view 3d
  perception.
\newblock {\em arXiv preprint arXiv:2303.05970}, 2023.

\bibitem{he2016resnet}
Kaiming He, Xiangyu Zhang, Shaoqing Ren, and Jian Sun.
\newblock Deep residual learning for image recognition.
\newblock In {\em CVPR}, 2016.

\bibitem{hu2023planning}
Yihan Hu, Jiazhi Yang, Li~Chen, Keyu Li, Chonghao Sima, Xizhou Zhu, Siqi Chai,
  Senyao Du, Tianwei Lin, Wenhai Wang, et~al.
\newblock Planning-oriented autonomous driving.
\newblock In {\em CVPR}, 2023.

\bibitem{huang2022bevdet4d}
Junjie Huang and Guan Huang.
\newblock {BEVDet4D}: Exploit temporal cues in multi-camera 3d object
  detection.
\newblock {\em arXiv preprint arXiv:2203.17054}, 2022.

\bibitem{huang2021bevdet}
Junjie Huang, Guan Huang, Zheng Zhu, and Dalong Du.
\newblock {BEVDet}: High-performance multi-camera 3d object detection in
  bird-eye-view.
\newblock {\em arXiv preprint arXiv:2112.11790}, 2021.

\bibitem{huang2023geometricaware}
Linyan Huang, Huijie Wang, Jia Zeng, Shengchuan Zhang, Liujuan Cao, Rongrong
  Ji, Junchi Yan, and Hongyang Li.
\newblock Geometric-aware pretraining for vision-centric 3d object detection.
\newblock {\em arXiv preprint arXiv:2304.03105}, 2023.

\bibitem{huang2022tigbev}
Peixiang Huang, Li~Liu, Renrui Zhang, Song Zhang, Xinli Xu, Baichao Wang, and
  Guoyi Liu.
\newblock Tig-bev: Multi-view bev 3d object detection via target inner-geometry
  learning.
\newblock {\em arXiv preprint arXiv:2212.13979}, 2022.

\bibitem{jia2023driveadapter}
Xiaosong Jia, Yulu Gao, Li~Chen, Junchi Yan, Patrick~Langechuan Liu, and
  Hongyang Li.
\newblock Driveadapter: Breaking the coupling barrier of perception and
  planning in end-to-end autonomous driving.
\newblock In {\em ICCV}, 2023.

\bibitem{jia2023thinktwice}
Xiaosong Jia, Penghao Wu, Li~Chen, Jiangwei Xie, Conghui He, Junchi Yan, and
  Hongyang Li.
\newblock Think twice before driving: Towards scalable decoders for end-to-end
  autonomous driving.
\newblock In {\em CVPR}, 2023.

\bibitem{jiang2022polarformer}
Yanqin Jiang, Li~Zhang, Zhenwei Miao, Xiatian Zhu, Jin Gao, Weiming Hu, and
  Yu-Gang Jiang.
\newblock {Polarforme}r: Multi-camera 3d object detection with polar
  transformers.
\newblock {\em arXiv preprint arXiv:2206.15398}, 2022.

\bibitem{klingner2023X3KD}
Marvin Klingner, Shubhankar Borse, Varun~Ravi Kumar, Behnaz Rezaei, Venkatraman
  Narayanan, Senthil Yogamani, and Fatih Porikli.
\newblock X$^3$kd: Knowledge distillation across modalities, tasks and stages
  for multi-camera 3d object detection.
\newblock {\em arXiv preprint arXiv:2303.02203}, 2023.

\bibitem{lang2019pointpillar}
Alex~H Lang, Sourabh Vora, Holger Caesar, Lubing Zhou, Jiong Yang, and Oscar
  Beijbom.
\newblock {PointPillars}: Fast encoders for object detection from point clouds.
\newblock In {\em CVPR}, 2019.

\bibitem{li2022delving}
Hongyang Li, Chonghao Sima, Jifeng Dai, Wenhai Wang, Lewei Lu, Huijie Wang,
  Enze Xie, Zhiqi Li, Hanming Deng, Hao Tian, et~al.
\newblock Delving into the devils of bird's-eye-view perception: A review,
  evaluation and recipe.
\newblock {\em arXiv preprint arXiv:2209.05324}, 2022.

\bibitem{li2022bevlgkd}
Jianing Li, Ming Lu, Jiaming Liu, Yandong Guo, Li~Du, and Shanghang Zhang.
\newblock Bev-lgkd: A unified lidar-guided knowledge distillation framework for
  bev 3d object detection.
\newblock {\em arXiv preprint arXiv:2212.00623}, 2022.

\bibitem{li2022uvtr}
Yanwei Li, Yilun Chen, Xiaojuan Qi, Zeming Li, Jian Sun, and Jiaya Jia.
\newblock Unifying voxel-based representation with transformer for 3d object
  detection.
\newblock {\em arXiv preprint arXiv:2206.00630}, 2022.

\bibitem{li2022bevstereo}
Yinhao Li, Han Bao, Zheng Ge, Jinrong Yang, Jianjian Sun, and Zeming Li.
\newblock Bevstereo: Enhancing depth estimation in multi-view 3d object
  detection with dynamic temporal stereo.
\newblock {\em arXiv preprint arXiv:2209.10248}, 2022.

\bibitem{li2022bevdepth}
Yinhao Li, Zheng Ge, Guanyi Yu, Jinrong Yang, Zengran Wang, Yukang Shi,
  Jianjian Sun, and Zeming Li.
\newblock {BEVDepth}: Acquisition of reliable depth for multi-view 3d object
  detection.
\newblock {\em arXiv preprint arXiv:2206.10092}, 2022.

\bibitem{li2022bevformer}
Zhiqi Li, Wenhai Wang, Hongyang Li, Enze Xie, Chonghao Sima, Tong Lu, Qiao Yu,
  and Jifeng Dai.
\newblock {BEVFormer}: Learning bird's-eye-view representation from
  multi-camera images via spatiotemporal transformers.
\newblock {\em arXiv preprint arXiv:2203.17270}, 2022.

\bibitem{li2023voxelformer}
Zhuoling Li, Chuanrui Zhang, Wei-Chiu Ma, Yipin Zhou, Linyan Huang, Haoqian
  Wang, SerNam Lim, and Hengshuang Zhao.
\newblock Voxelformer: Bird's-eye-view feature generation based on dual-view
  attention for multi-view 3d object detection.
\newblock {\em arXiv preprint arXiv:2304.01054}, 2023.

\bibitem{liang2022bevfusion}
Tingting Liang, Hongwei Xie, Kaicheng Yu, Zhongyu Xia, Zhiwei Lin, Yongtao
  Wang, Tao Tang, Bing Wang, and Zhi Tang.
\newblock {BEVFusion}: A simple and robust lidar-camera fusion framework.
\newblock {\em arXiv preprint arXiv:2205.13790}, 2022.

\bibitem{liu2022petr}
Yingfei Liu, Tiancai Wang, Xiangyu Zhang, and Jian Sun.
\newblock {PETR}: Position embedding transformation for multi-view 3d object
  detection.
\newblock {\em arXiv preprint arXiv:2203.05625}, 2022.

\bibitem{liu2022petrv2}
Yingfei Liu, Junjie Yan, Fan Jia, Shuailin Li, Qi~Gao, Tiancai Wang, Xiangyu
  Zhang, and Jian Sun.
\newblock {PETRv2}: A unified framework for 3d perception from multi-camera
  images.
\newblock {\em arXiv preprint arXiv:2206.01256}, 2022.

\bibitem{liu2021swin}
Ze~Liu, Yutong Lin, Yue Cao, Han Hu, Yixuan Wei, Zheng Zhang, Stephen Lin, and
  Baining Guo.
\newblock Swin transformer: Hierarchical vision transformer using shifted
  windows.
\newblock In {\em ICCV}, 2021.

\bibitem{liu2022convnet}
Zhuang Liu, Hanzi Mao, Chao-Yuan Wu, Christoph Feichtenhofer, Trevor Darrell,
  and Saining Xie.
\newblock A convnet for the 2020s.
\newblock In {\em CVPR}, 2022.

\bibitem{park2022solofusion}
Jinhyung Park, Chenfeng Xu, Shijia Yang, Kurt Keutzer, Kris Kitani, Masayoshi
  Tomizuka, and Wei Zhan.
\newblock Time will tell: New outlooks and a baseline for temporal multi-view
  3d object detection.
\newblock {\em arXiv preprint arXiv:2210.02443}, 2022.

\bibitem{romero2014fitnets}
Adriana Romero, Nicolas Ballas, Samira~Ebrahimi Kahou, Antoine Chassang, Carlo
  Gatta, and Yoshua Bengio.
\newblock Fitnets: Hints for thin deep nets.
\newblock {\em arXiv preprint arXiv:1412.6550}, 2014.

\bibitem{shi2021pvrcnn++}
Shaoshuai Shi, Li~Jiang, Jiajun Deng, Zhe Wang, Chaoxu Guo, Jianping Shi,
  Xiaogang Wang, and Hongsheng Li.
\newblock {PV-RCNN++}: Point-voxel feature set abstraction with local vector
  representation for 3d object detection.
\newblock {\em arXiv preprint arXiv:2102.00463}, 2021.

\bibitem{shu2021cwd}
Changyong Shu, Yifan Liu, Jianfei Gao, Zheng Yan, and Chunhua Shen.
\newblock Channel-wise knowledge distillation for dense prediction.
\newblock In {\em ICCV}, 2021.

\bibitem{Tong_2023_ICCV}
Wenwen Tong, Chonghao Sima, Tai Wang, Li~Chen, Silei Wu, Hanming Deng, Yi~Gu,
  Lewei Lu, Ping Luo, Dahua Lin, and Hongyang Li.
\newblock Scene as occupancy.
\newblock In {\em ICCV}, 2023.

\bibitem{vaswani2017attention}
Ashish Vaswani, Noam Shazeer, Niki Parmar, Jakob Uszkoreit, Llion Jones,
  Aidan~N Gomez, {\L}ukasz Kaiser, and Illia Polosukhin.
\newblock Attention is all you need.
\newblock In {\em NeurIPS}, 2017.

\bibitem{vora2020pointpainting}
Sourabh Vora, Alex~H Lang, Bassam Helou, and Oscar Beijbom.
\newblock {PointPainting}: Sequential fusion for 3d object detection.
\newblock In {\em CVPR}, 2020.

\bibitem{wang2023openlanev2}
Huijie Wang, Tianyu Li, Yang Li, Li~Chen, Chonghao Sima, Zhenbo Liu, Yuting
  Wang, Shengyin Jiang, Peijin Jia, Bangjun Wang, Feng Wen, Hang Xu, Ping Luo,
  Junchi Yan, Wei Zhang, and Hongyang Li.
\newblock Openlane-v2: A topology reasoning benchmark for scene understanding
  in autonomous driving, 2023.

\bibitem{wang2023streampetr}
Shihao Wang, Yingfei Liu, Tiancai Wang, Ying Li, and Xiangyu Zhang.
\newblock Exploring object-centric temporal modeling for efficient multi-view
  3d object detection.
\newblock {\em arXiv preprint arXiv:2303.11926}, 2023.

\bibitem{wang2021fcos3d}
Tai Wang, Xinge Zhu, Jiangmiao Pang, and Dahua Lin.
\newblock {FCOS3D}: Fully convolutional one-stage monocular 3d object
  detection.
\newblock In {\em ICCV}, 2021.

\bibitem{wang2022detr3d}
Yue Wang, Vitor~Campagnolo Guizilini, Tianyuan Zhang, Yilun Wang, Hang Zhao,
  and Justin Solomon.
\newblock {DETR3D}: 3d object detection from multi-view images via 3d-to-2d
  queries.
\newblock In {\em CoRL}, 2022.

\bibitem{wang2022sts}
Zengran Wang, Chen Min, Zheng Ge, Yinhao Li, Zeming Li, Hongyu Yang, and
  Di~Huang.
\newblock {STS}: Surround-view temporal stereo for multi-view 3d detection.
\newblock {\em arXiv preprint arXiv:2208.10145}, 2022.

\bibitem{wu2022trajectoryguided}
Penghao Wu, Xiaosong Jia, Li~Chen, Junchi Yan, Hongyang Li, and Yu~Qiao.
\newblock Trajectory-guided control prediction for end-to-end autonomous
  driving: A simple yet strong baseline.
\newblock In {\em Advances in Neural Information Processing Systems (NeurIPS)},
  2022.

\bibitem{Xiong_2023_CVPR}
Kaixin Xiong, Shi Gong, Xiaoqing Ye, Xiao Tan, Ji~Wan, Errui Ding, Jingdong
  Wang, and Xiang Bai.
\newblock Cape: Camera view position embedding for multi-view 3d object
  detection.
\newblock In {\em CVPR}, pages 21570--21579, June 2023.

\bibitem{yan2023cmt}
Junjie Yan, Yingfei Liu, Jianjian Sun, Fan Jia, Shuailin Li, Tiancai Wang, and
  Xiangyu Zhang.
\newblock Cross modal transformer via coordinates encoding for 3d object
  dectection.
\newblock {\em arXiv preprint arXiv:2301.01283}, 2023.

\bibitem{yang2022bevformer}
Chenyu Yang, Yuntao Chen, Hao Tian, Chenxin Tao, Xizhou Zhu, Zhaoxiang Zhang,
  Gao Huang, Hongyang Li, Yu~Qiao, Lewei Lu, et~al.
\newblock Bevformer v2: Adapting modern image backbones to bird's-eye-view
  recognition via perspective supervision.
\newblock {\em arXiv preprint arXiv:2211.10439}, 2022.

\bibitem{yang2022deepinteraction}
Zeyu Yang, Jiaqi Chen, Zhenwei Miao, Wei Li, Xiatian Zhu, and Li~Zhang.
\newblock {DeepInteraction}: 3d object detection via modality interaction.
\newblock {\em arXiv preprint arXiv:2208.11112}, 2022.

\bibitem{yin2021centerpoint}
Tianwei Yin, Xingyi Zhou, and Philipp Krahenbuhl.
\newblock Center-based 3d object detection and tracking.
\newblock In {\em CVPR}, 2021.

\bibitem{Zeng_2023_CVPR}
Jia Zeng, Li~Chen, Hanming Deng, Lewei Lu, Junchi Yan, Yu~Qiao, and Hongyang
  Li.
\newblock Distilling focal knowledge from imperfect expert for 3d object
  detection.
\newblock In {\em CVPR}, 2023.

\bibitem{zhao2023bevsimdet}
Haimei Zhao, Qiming Zhang, Shanshan Zhao, Jing Zhang, and Dacheng Tao.
\newblock Bevsimdet: Simulated multi-modal distillation in bird's-eye view for
  multi-view 3d object detection.
\newblock {\em arXiv preprint arXiv:2303.16818}, 2023.

\bibitem{zhou2019center}
Xingyi Zhou, Dequan Wang, and Philipp Kr{\"a}henb{\"u}hl.
\newblock Objects as points.
\newblock {\em arXiv preprint arXiv:1904.07850}, 2019.

\bibitem{zhou2021monocular}
Yunsong Zhou, Yuan He, Hongzi Zhu, Cheng Wang, Hongyang Li, and Qinhong Jiang.
\newblock Monocular 3d object detection: An extrinsic parameter free approach.
\newblock In {\em CVPR}, 2021.

\bibitem{Zhou2022monocular}
Yunsong Zhou, Yuan He, Hongzi Zhu, Cheng Wang, Hongyang Li, and Qinhong Jiang.
\newblock Monoef: Extrinsic parameter free monocular 3d object detection.
\newblock {\em IEEE TPAMI}, 2022.

\end{thebibliography}
\clearpage

\appendix


\section{Experiment Details}
\label{sec:experiments}

\subsection{Dataset and Evaluation Metrics}
We conduct our experiments on the nuScenes dataset~\cite{caesar2020nuscenes}, a widely used benchmark for autonomous driving tasks. The dataset encompasses diverse driving scenarios captured using cameras and LiDAR sensors, offering rich information for both visual and LiDAR-based 3D object detection. The dataset comprises 700 training scenes, 150 validation scenes, and 150 testing scenes. Each scene spans approximately 20 seconds, with key frames annotated at a 2 Hz frequency.

The two dominant metrics for the nuScenes detection task are the nuScenes Detection Score (NDS) and mean Average Precision (mAP). The mAP for nuScenes is computed based on the center distance between predictions and ground truth annotations on the ground plane. Moreover, the nuScenes dataset defines five true positive metrics (mATE, mASE, mAOE, mAVE, mAAE) for measuring translation, scale, orientation, velocity, and attribute, respectively. The NDS for nuScenes is a weighted sum of mAP and the five true positive metrics, defined as $NDS = \frac{1}{10}[5mAP+\sum_{mTP}(1-\min(1, mTP))]$.

\subsection{Implementation Details}
We conduct experiments on BEVDepth~\cite{li2022bevdepth}. 
The codebase is developed upon MMDetection3D~\cite{mmdet3d2020}. Main experiments are trained on 8 NVIDIA A100 GPUs, while ablation experiments are conducted on 8 NVIDIA V100 GPUS. For BEVDepth, the model is trained for 20 epochs with an initial learning rate of 2e-4. In the distillation process, the per-GPU batch size is set to 4, whereas during the training of the baseline model, it is set to 8. Normal data augmentations are introduced in the training process such as flip and rotate. In our apprentice models, future frames are not incorporated into the long-term temporal fusion throughout the training phase to ensure a fair comparison. As for ablation study, we conduct experiments with an online training strategy, and we have employed the ResNet-50 configuration in the absence of CBGS.

In our research, we implement distinct temporal modeling strategies for both apprentice and expert models. For the apprentice models, we incorporate a sequence of eight frames into the temporal modeling process. In contrast, the expert models integrate four future frames into the temporal modeling as demonstrated in our primary results. However, in our ablation study, we deviate from this approach and instead employ eight historical frames for temporal modeling. Besides, We use the last sweep LiDAR frame with the current LiDAR frame to create the depth map.

\begin{table}[hb]
  \caption{Experiment settings. $^*$ denotes that the training schedule for VCD-E is approximately one-fourth of the original schedule. This reduction was implemented to expedite the training process during the ablation study. The first group is engaged in training on the main results, whereas the second group is utilized in the ablation study.
  }
  \label{settings}
  \centering
  \begin{tabular}{{c|c|c|c|c|c}}
  \toprule
  Method   & Backbone & Image Size & Frames &  mAP (\%) & NDS (\%) \\ 
  \midrule
  VCD-E  & ConvNext-B~\cite{liu2022convnet}  & 512 $\times$ 1408 & 8+1 & 67.7 & 71.1 \\ 
  VCD-A  & Res-50~\cite{he2016resnet}  & 256 $\times$ 704 & 8+1 & 41.8 & 54.2 \\ 
  \midrule
   VCD-E$^*$ & ConvNext-B~\cite{liu2022convnet} & 256 $\times$ 704 & 8+1 & 54.2 & 58.8 \\ 
   VCD-A & Res-50~\cite{he2016resnet}  &  256 $\times$ 704 & 8+1 & 29.7  & 40.9 \\ 
  \bottomrule
  \end{tabular}
\end{table}

\subsection{Experiments Settings}

The setting of adopted expert-apprentice pairs is depicted in Tab.~\ref{settings}. We categorize the distillation setting into two distinct groups. The primary group is engaged in training on the main results, whereas the second group is utilized for the ablation study. Both our baseline and VCD-A adopt 8 history frames for temporal fusion.

\section{The Analysis of Temporal Fusion}
\label{sec:temporal}

\subsection{The Misalignment of Motion Objects}

As highlighted in preceding studies~\cite{park2022solofusion}, long-term temporal fusion may face misalignment issues in motion estimation, which can be discerned through a reduction in performance on metrics like mATE. Let's consider a moving object and analyze the impact of inaccurate motion estimation on its position in the fused frame. We will assume that the environment is static, except for the moving object. Let the position of the moving object in the world coordinate system be represented by $\boldsymbol{P}_i^w = (x_i^w, y_i^w, z_i^w, 1)^T$ in each of the $N$ frames captured at times $t_1, t_2, \dots, t_N$. The actual motion of the moving object between frames is represented by $\boldsymbol{M}_i^{obj}$, and the estimated motion is represented by $\boldsymbol{\hat{M}}_i^{obj}$. The difference between the estimated and actual motion of the object can be denoted as:
\begin{equation}
\Delta \boldsymbol{M_i}^{obj} = \boldsymbol{M_i}^{obj} - \boldsymbol{\hat{M}}_i^{obj}. 
\end{equation}

As we have already computed the transformation matrix $\boldsymbol{T}_i$ based on the estimated ego motion, we can calculate the transformed object position in the current frame, considering its actual motion, as:
\begin{equation}
\boldsymbol{P}_i^{w'} = \boldsymbol{T}_i \times \boldsymbol{M}_i^{obj} \times\boldsymbol{P}_i^w. 
\end{equation}

The error in the transformed object position can be computed as:
\begin{equation}
\boldsymbol{e}_i^{obj} = \boldsymbol{P}_i^{w'} - \boldsymbol{\hat{P}}_i^{w'}. 
\end{equation}

In the long-term fusion process, we integrate the information from all $N$ frames. Assuming we use a fusion function $F$, the fused position in the current frame can be represented as:
\begin{equation}
\boldsymbol{P}_{fusion}^{obj} = F(\boldsymbol{P}_1^{w'}, \boldsymbol{P}_2^{w'}. \dots, \boldsymbol{P}_N^{w'}).
\end{equation}

The inaccuracies in the motion estimation of the moving object for each frame can propagate through the fusion function and result in a misaligned object in the fused frame. The overall error in the fused position can be represented as a function of the errors in each frame:
\begin{equation}
\boldsymbol{e}_{fusion}^{obj} = G(\boldsymbol{e}_1^{obj}, \boldsymbol{e}_2^{obj}. \dots, \boldsymbol{e}_N^{obj}),
\end{equation}
where $G$ represents a function that combines the errors from each frame. The fused position of the moving object will be less accurate due to these motion estimation errors, leading to a decline in object detection performance in the long-term setting. To address the issue mentioned earlier, we introduce the trajectory-based distillation module, which compensates for the misalignment of moving objects. We will provide further details in the subsequent discussion.

\begin{figure}[tb]
  \centering
  \includegraphics[width=\linewidth]{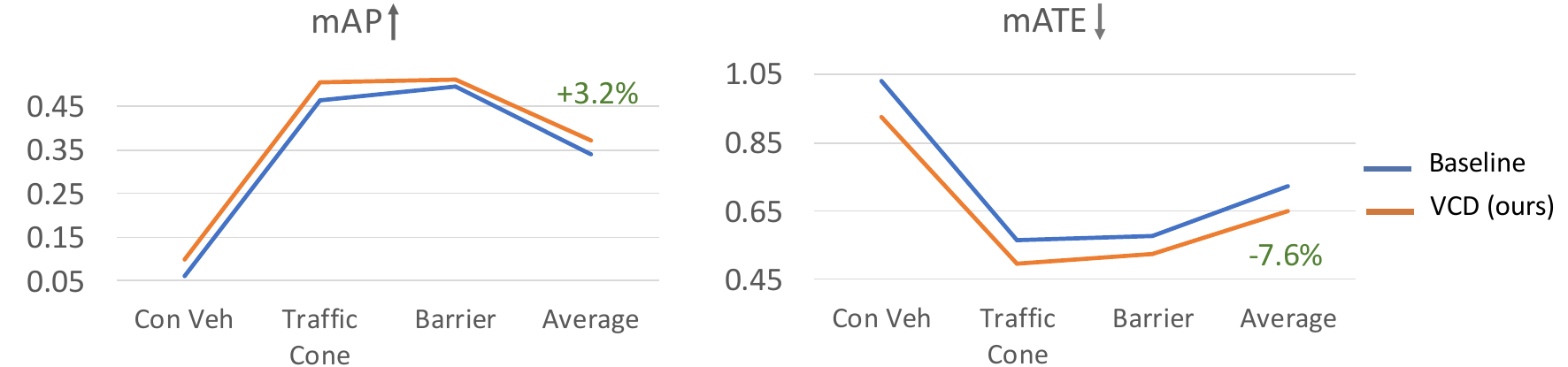}
  \caption{Effects of VCD on static objects. Our distillation framework VCD can still consistently improves static objects, demonstrating 3.2\% and 7.6\% improvements in precision-recall (mAP) and localization quality (mATE), respectively. }
  \label{analysis3}
\end{figure}

\begin{figure}
  \centering
  \includegraphics[width=.95\linewidth]{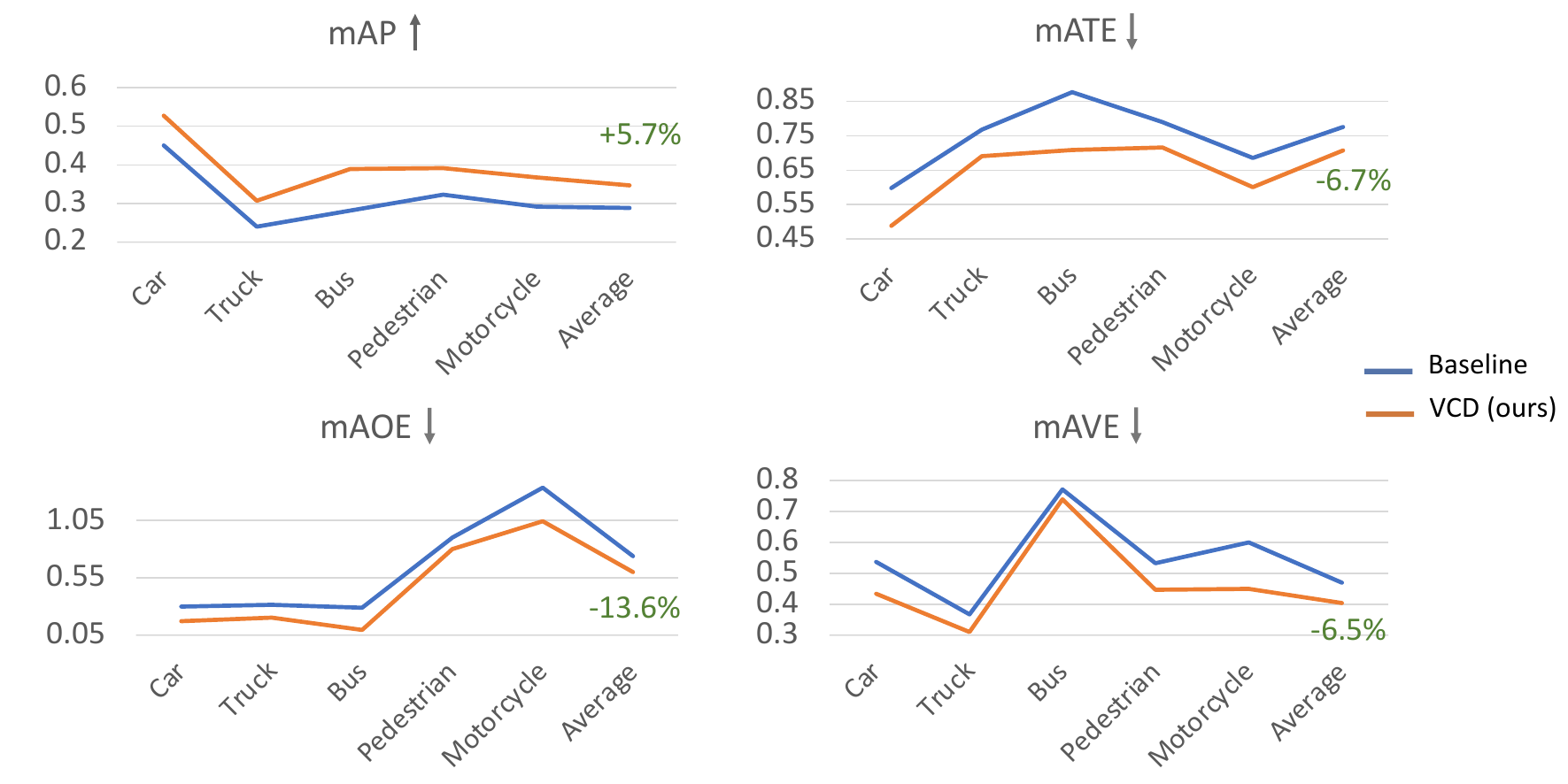}
  \caption{Effects of VCD on movable objects. Our distillation framework VCD consistently improves dynamic objects across a range of metrics.}
  \label{analysis}
\end{figure}

\subsection{The Improvements of Static and Dynamic Objects}

In this section, we present visualizations to demonstrate the improvements achieved in dynamic objects. Particularly noteworthy is the significant enhancement in the representation of dynamic objects through trajectory-based distillation, thereby highlighting the effectiveness of the trajectory-based module. As depicted in Fig.~\ref{analysis3} and Fig.~\ref{analysis}, our distillation framework consistently enhances static and dynamic objects across various metrics.




\section{Visualization}
\label{sec:visualization}

\begin{figure}
  \centering
  \includegraphics[width=.99\linewidth]{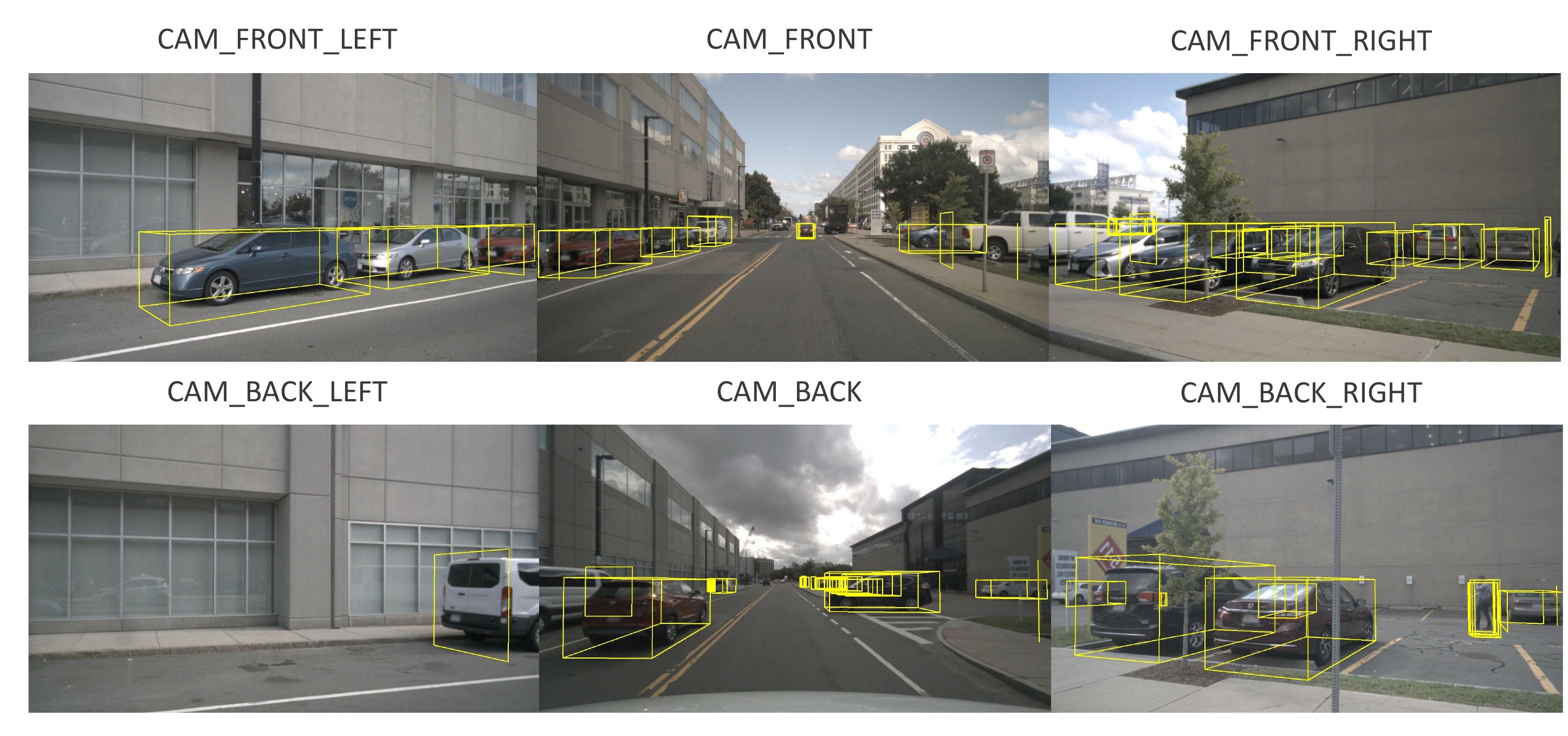}
  \caption{Visualization of the predictions for 3D object detection generated by the VCD-A.}
  \label{visulization}
\end{figure}

We have performed several visualizations in Fig.~\ref{visulization} to showcase the advancements achieved by our distillation framework. Our findings indicate that our models excel in accurately predicting 3D bounding boxes for the target objects.

\section{Broader Impact}
\label{sec:impact}
Our research introduces a novel perspective for multi-modal methodologies and a fresh distillation paradigm for camera-only techniques. We believe that it can establish a robust baseline for the broader scientific community. However, while our methods contribute to the enhancement of autonomous driving, they are not yet capable of addressing more complex corner cases. Consequently, these limitations could potentially introduce risks in real-world autonomous systems.

\end{document}